\documentclass[runningheads]{llncs}

\usepackage{eccv}

\usepackage{eccvabbrv}

\usepackage{graphicx}
\usepackage{booktabs}
\usepackage{subcaption}
\usepackage[accsupp]{axessibility}  
\usepackage{algorithm}
\usepackage{algpseudocode}
\usepackage{wrapfig}
\usepackage{placeins}
\usepackage[utf8]{inputenc} 
\DeclareUnicodeCharacter{200E}{}
\usepackage{colortbl}%
  \newcommand{\myrowcolor}{\rowcolor[gray]{0.925}}

\usepackage[pagebackref,breaklinks,colorlinks]{hyperref}

\usepackage{orcidlink}

\begin{document}

\title{Tackling Structural Hallucination in Image Translation with Local Diffusion} 


\author{%
{Seunghoi Kim$^*$}\inst{1,2,4}\orcidlink{0009-0009-8456-8448} \and
{Chen Jin$^*$}\inst{4}\orcidlink{0000-0002-2179-6445} \and 
{Tom Diethe}\inst{4} \and
{Matteo Figini}\inst{1,3} \and
{Henry F. J. Tregidgo}\inst{1,2} \and
{Asher Mullokandov}\inst{4} \and
{Philip Teare}\inst{4} \and
{Daniel C. Alexander}\inst{1,3}\orcidlink{0000-0003-2439-350X}
}

\authorrunning{Kim et al.}

\institute{
Centre for Medical Image Computing, University College London \and Department of Medical Physics and Biomedical Engineering, University College London \and Department of Computer Science, University College London \and Centre for AI, DS\&AI, AstraZeneca, UK
}

\maketitle
\setcounter{footnote}{0}
\def\thefootnote{*}\footnotetext{These authors contributed equally to this work \\
Corresponding author E-mail: seunghoi.kim.17@ucl.ac.uk}\def\thefootnote{\arabic{footnote}}
\begin{abstract}
Recent developments in diffusion models have advanced conditioned image generation, yet they struggle with reconstructing out-of-distribution (OOD) images, such as unseen tumors in medical images, causing ``image hallucination'' and risking misdiagnosis. We hypothesize such hallucinations result from local OOD regions in the conditional images. 
We verify that partitioning the OOD region and conducting separate image generations alleviates hallucinations in several applications. From this, we propose a training-free diffusion framework that reduces hallucination with multiple \textit{Local Diffusion} processes. Our approach involves OOD estimation followed by two modules: a ``branching'' module generates locally both within and outside OOD regions, and a ``fusion'' module integrates these predictions into one. Our evaluation shows our method mitigates hallucination over baseline models quantitatively and qualitatively, reducing misdiagnosis by 40\% and 25\% in the real-world medical and natural image datasets, respectively. It also demonstrates compatibility with various pre-trained diffusion models. Code is available at \url{https://github.com/edshkim98/LocalDiffusion-Hallucination}.
\end{abstract}

\section{Introduction}
\label{sec:intro}
\begin{figure}[H]
  \centering
  \includegraphics[width=1\textwidth]{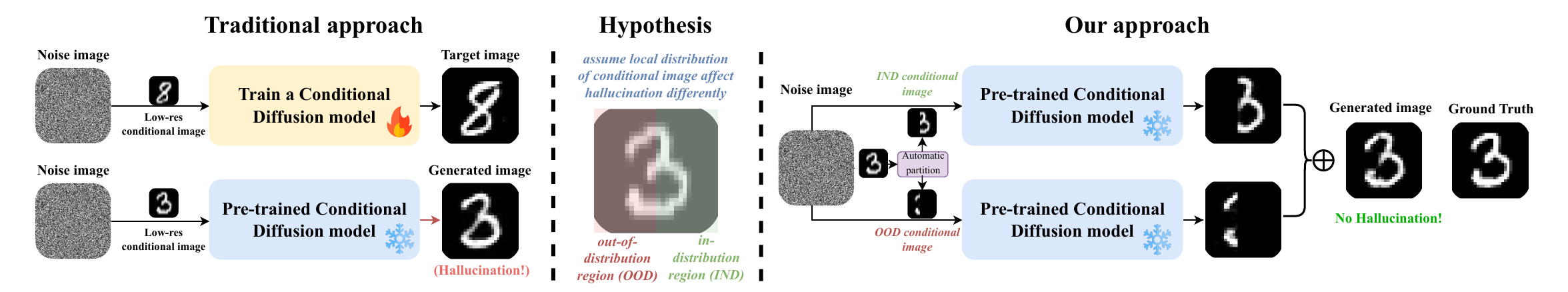}
  \caption{Traditional conditional diffusion process vs. our OOD/IND \textit{Local Diffusion}.}
  \label{fig:intro_schematic}
\end{figure}

\begin{figure}[H]
\centering
\includegraphics[width=1\textwidth]{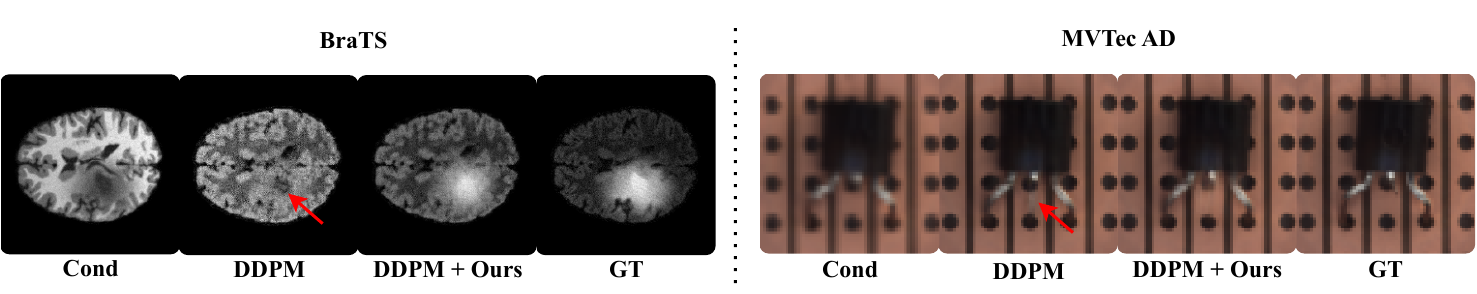}
\caption{Examples of structural hallucinations (indicated by arrows) in DDPM\cite{ddpm} compared with the outcomes of our method for BraTS image translation (left) and MVTec super-resolution (right).}\label{fig:intro_hallucinations}
\end{figure}

Conditional diffusion models\cite{ddpm, ddim, diffusion_improved, variational_diffusion, stablediffusion, imagen} have attained state-of-the-art performance across a range of image translation tasks, such as colorization \cite{sdedit}, enhancement \cite{diffusioniqt, srdiff}, and denoising \cite{ddrm}. These tasks typically necessitate a conditional image (e.g., low-resolution or grayscale) as input to produce realistic and accurate outputs (see examples in Fig~\ref{fig:intro_schematic} and Fig~\ref{fig:intro_hallucinations}).

\vspace{-4mm} 
While diffusion models excel in generation tasks using conditional images within their training distribution, they encounter difficulties with out-of-distribution (OOD) features. These models often generate predictions that diverge from the intended output when processing data beyond their training scope, despite their improved generalization capabilities over earlier deep learning methods\cite{score_mri, diffusioniqt}.
\vspace{-2mm} 

\vspace{-4mm} 

As shown in the left example in Fig~\ref{fig:intro_hallucinations}, DDPM\cite{ddpm} trained on healthy brain images failed to recognize and accurately process OOD elements (tumors). This misinterpretation can mistakenly transform a tumor into healthy tissue, potentially posing serious risks in medical diagnostics. We term this issue \textit{``structural hallucination''}, where models generate realistic-looking but inaccurate reconstructions, leading to discrepancies with the actual structure. This is critical in medical imaging and applications with significant under-representation of classes, which are yet under-explored.

Aiming to train or fine-tune models to include all OOD datasets can be impractical or require significant resources.
Alternative methods, including zero-shot \cite{dip, zeroddrm, patchOOD} and distribution-shift inversion techniques (DSI)\cite{inversionshift}, have been proposed. Yet, such approaches encode prior knowledge of test input rather than external training data and generally introduce additional inference time. While DSI does not demand re-training of the pre-trained diffusion model, its effectiveness is primarily confined to style transfer tasks. In-context learning\cite{dreambooth, suti}, which utilizes pre-trained models to adapt to new tasks without extensive training, but they are mainly constrained to natural language processing.

In this paper, we focus on reducing structural hallucination. We hypothesize that \textit{hallucinations are caused by local OOD regions in the conditional images, and by partitioning the OOD area from in-distribution (IND) region and conducting separate generations, hallucinations can be alleviated}.
We verified our hypothesis by conducting motivational experiments involving the manual partitioning of these regions and their independent generation. We found out that separate generations of OOD/IND regions reduce structural hallucination and the generation of such hallucinations is prone in the early/mid stage of the diffusion process.
Building on these insights, we propose a novel diffusion process aimed at reducing the hallucination in pre-trained diffusion models without any additional training with new data. To the best of our knowledge, this is the first work to identify and tackle the hallucination problem in diffusion models for image translation.
Our method initially utilizes an anomaly detector to estimate a probabilistic OOD map of the conditional image and create two masked OOD/IND conditional images. Then, our diffusion model implements a two-step procedure. First, a ``branching'' module, which performs two local generations in parallel, corresponding to IND and OOD regions, respectively. Then a ``fusion'' module merges the two local generations, which minimizes the creation of hallucinated features. This technique does not necessitate fine-tuning, offering a cost-effective way to mitigate structural hallucination, and can plug-and-play into any pre-trained diffusion models. Our contributions can be summarized as follows:

\begin{itemize}
  \item We demonstrate the impact of out-of-distribution conditional image on image generation using diffusion models, and show from motivational experiments that local image generation based on OOD/IND segmentation can reduce structural hallucination.
  \item We identify hallucination hotspots in diffusion models, occurring predominately in the early and middle stages of the reverse process. 
  \item We propose a novel framework that performs \textit{Local Diffusion} processes in parallel to minimize structural hallucination, without the need for re-training the pre-trained diffusion model.
  \item We demonstrate that our approach reduces image hallucination and leads to significantly reduced misdiagnosis risks by 40\% in the medical lesion segmentation and 25\% in natural image anomaly detection tasks.
\end{itemize}

\section{Related Work}
\subsection{Diffusion Models}

Diffusion models \cite{ddpm,ddim,diffusion_improved, diff_survey}, a class of generative models, have recently emerged as a leading paradigm in image generation. Outperforming GANs\cite{stylegan, biggan, agcn} in image quality and demonstrating greater training stability\cite{dmbeatsgans, diffusiongan}, diffusion models are gaining extensive attention in areas such as image super-resolution\cite{srdiff, diffusioniqt}, inpainting\cite{repaint,controlnet}, and enhancement \cite{ddrm, zeroddrm, improving_inverse}. These models employ a neural network that incrementally denoises an image across multiple time steps. To address issues with extensive sampling requirements, various approaches have been proposed, including deterministic non-Markovian reverse processes\cite{ddim}, knowledge distillation\cite{diffusion_kd, consistency}, and adversarial learning\cite{diffusion_adv} .

\subsection{Out-of-Distribution Generalization} 

Deep learning models often struggle with generalizing to OOD data, leading to hallucinated features. Initially, strategies such as Deep Image Prior \cite{dip} attempted to learn from measured data, requiring test-time training which extended inference time and typically falling short of supervised models due to the lack of leveraging pre-trained knowledge.

Recent advancements have aimed to enhance diffusion models' OOD performance. Text-to-image models\cite{dreambooth, textualinversion} have seen fine-tuning with limited datasets and in-context learning methods \cite{suti, instructpix2pix} to bypass fine-tuning. The Distribution Shift Inversion (DSI) algorithm \cite{inversionshift} presents a fine-tuning free method by mixing test samples with Gaussian noise to align them closer to the training distribution, showing promise mainly in style transfer task. Patch-based method \cite{patchOOD} proposes a local auto-regressive model that exclusively models local image features to improve OOD performance. 

However, most prior work has prioritized improving overall image quality metrics over mitigating hallucinations. As a fine-tuning free approach, our work addresses reducing structural hallucinations as well as improving image quality across image-to-image translation tasks from pre-trained diffusion models.

\section{Methodology}
\subsection{Preliminaries: Conditional Denoising Diffusion Probabilistic Models}
We build upon a denoising diffusion probabilistic model (DDPM)\cite{ddpm} to develop our conditional diffusion model, incorporating an encoder for conditional inputs that merges with the attention U-Net at the bottleneck, trained simultaneously.

For a given clean image $x$ and its noisy counterpart $x_{t}$ at any time step $t$ (where $0 \leq t \leq T$), the forward process incrementally adds Gaussian noise to $x$, culminating in isotropic Gaussian noise at $t=T$. This process is described by:
\begin{equation}
\label{eq:forward}
q(x_{t}|x_{t-1}) = \mathcal{N}(x_{t}, \sqrt{1-\beta_{t}}x_{t-1}, \beta_{t}I) = \mathcal{N}(x_{t}, \sqrt{\bar{\alpha}_{t}}x, (1-\bar{\alpha}_{t})I) 
\end{equation}
where $\alpha_{t} = 1 - \beta_{t}$, $\bar{\alpha}_{t}=\prod_{i=1}^{t}\alpha_{s}$, and $\beta_{t}$ the noise scheduler dictates the addition of Gaussian noise $\epsilon$ using the sigmoid function.

During the reverse process, the model reconstructs $x$ from isotropic Gaussian noise ($t=T$), by gradually denoising it. We enhance this process by conditioning an additional image, such as a low-resolution version, at each step.

The variational lower bound\cite{ddpm} leads to multiple parameterizations. If $\hat{x}_{\theta}$ symbolizes a neural network predicting $x$ from a noisy image $x_{t}$ and a condition image $x_{c}$, the loss function $L$ measures the difference between $x$ and $\hat{x}_{\theta}$'s output:
\begin{equation}
\label{eq:continuous_loss}
L =  \mathbb{E}_{x,t,\varepsilon}[||x - \hat{x}_{\theta}(\alpha_{t}x + \sigma_{t}\varepsilon,x_{c}, t)||^2],
\end{equation}
with $\alpha_{t}$ and $\sigma_{t}$ as time-dependent scalars from the noise schedule.

\subsection{Hypothesis and Justification}

Assume that both $x_{t-1}$ and $x_{t}$ can be partitioned into in-distribution and out-of-distribution components, denoted as $x_{t-1}^{in}$, $x_{t-1}^{out}$ and $x_{t}^{in}$, $x_{t}^{out}$ respectively. Then, in estimating $x_{t-1}^{in}$, the term $x_{t}^{in}$ contains ``useful'' information, while $x_{t}^{out}$ is considered to contain ``noise'' or information less relevant to the distribution in question. The same rationale applies to $x_{t-1}^{out}$. We hypothesize that the conventional denoising process, which estimates $p(x_{t-1}|x_{t})$, is conditioned on partially noised data and is, therefore, imprecise. By conditioning on more refined information that distinctly recognizes in-distribution from out-of-distribution data, we posit that the Evidence Lower Bound (ELBO) with partitioning can be equal to or exceed the standard ELBO. This is due to a potentially reduced Kullback-Leibler (KL) divergence over the time steps, which we substantiate as follows.

In DDPM, the ELBO without partition is given by:
\begin{equation}
\begin{gathered}
ELBO = \mathbb{E}_{q}[\log p(x_0) - KL(q(x_{t-1}|x_{t}) \| p(x_{t-1}|x_{t}))]
\end{gathered}
\end{equation}
Let's consider the $x_{t-1}$ at each reverse step can be partitioned into $x_{t-1}^{in}$ and $x_{t-1}^{out}$, but both conditioned to the un-partitioned $x_{t}$. This is equivalent to the original reverse process and the $ELBO$ can be rewritten as $ELBO_{partition}^{est}$:
\begin{equation}
\begin{gathered}
ELBO_{partition}^{est} = ELBO_{in}^{est} + ELBO_{out}^{est} = \\
\mathbb{E}_{q}[\log p(x_0) - KL(q(x_{t-1}^{in}|x_{t}) 
\| p(x_{t-1}^{in}|x_{t}))] + \\
\mathbb{E}_{q}[\log p(x_0) - 
KL(q(x_{t-1}^{out}|x_{t}) \| p(x_{t-1}^{out}|x_{t}))]
\end{gathered}
\end{equation}

Now, let's consider both estimation $x_{t-1}$ and condition $x_{t}$ are partitioned, then the $ELBO_{partition}^{est\_cond}$:
\begin{equation}
\begin{gathered}
ELBO_{partition}^{est\_cond} = ELBO_{in}^{est\_cond} + ELBO_{out}^{est\_cond} = \\
\mathbb{E}_{q}[\log p(x_0^{in}) - KL(q(x_{t-1}^{in}|x_{t}^{in}) 
\| p(x_{t-1}^{in}|x_{t}^{in}))] + \\
\mathbb{E}_{q}[\log p(x_0^{out}) - 
KL(q(x_{t-1}^{out}|x_{t}^{out}) \| p(x_{t-1}^{out}|x_{t}^{out}))]
\end{gathered}
\end{equation}

The key difference between these $ELBO_{partition}^{est}$ and $ELBO_{partition}^{est\_cond}$ is in the KL divergence terms. With $ELBO_{partition}^{est}$, the $ELBO_{in}^{est}$ and $ELBO_{out}^{est}$ both conditioned on partially noised data. While $ELBO_{partition}^{est\_cond}$ allows the model to specifically account for the separate contributions of in-distribution and out-of-distribution data, hence we know
$ELBO_{in}^{est} \leq ELBO_{in}^{est\_cond}, ELBO_{out}^{est} \leq ELBO_{out}^{est\_cond}$. Therefore we can justify:

\begin{equation}
    ELBO = ELBO_{in}^{est} \leq ELBO_{partition}^{est\_cond}
\end{equation}

Hence partitioning both estimation and condition data into in-distribution and out-of-distribution and performing separate denoising processes provides a more informed and hence likely tighter bound on the reverse process, leading to a potentially higher (or at worst, equal) ELBO.

\subsection{Motivational Experiments: Hypothesis Verification}
This section presents experiments aimed at validating our hypothesis through MNIST\cite{mnist} super-resolution and BraTS\cite{brats} image translation (from T1 to Falir). In this experiment, our conditional DDPM was trained exclusively on the MNIST digit `8' or healthy brain images from BraTS. For testing, it is challenged with the MNIST digit `3' or BraTS images featuring tumors to specifically assess hallucinations in the generated images. 

\subsubsection{Can OOD-based Local Image Generation Help to Reduce Hallucination?}
To address hallucinations, we distinguished between IND and OOD regions in conditional images using segmentation masks, as shown in Fig~\ref{fig:intro_schematic}. These regions were manually separated, drawing on a deep understanding of the dataset. During inference, we applied a distinct reverse process for each sub-image, effectively separating each region in both the condition and the prediction. Combining predictions from both regions subsequently formed a complete image. Our approach, contrasted with traditional DDPM reverse process as shown in Fig~\ref{fig:intro_schematic}, notably reduced structural hallucinations. By focusing on each region independently, we demonstrated that our segmentation strategy effectively reduces such artefacts, confirming its utility in enhancing image generation fidelity.

\subsubsection{The Impact of OOD Segmentation on Hallucinations}
\begin{figure}[H]
\centering
\includegraphics[width=1.0\textwidth]{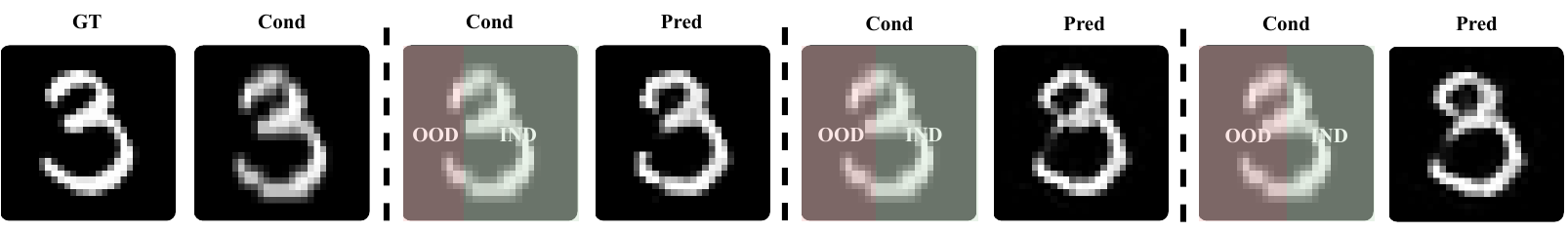}
\caption{Impact of different OOD boundaries on predictions.}\label{fig:motive2}
\end{figure}

To demonstrate the necessity of precise OOD segmentation for reducing hallucinations, we varied the boundary delineation between OOD and IND areas. 
Fig~\ref{fig:motive2} showcases the outcomes of employing different boundary indices. Notably, minor boundary adjustments resulted in significant predictive discrepancies, underscoring the importance of accurate OOD segmentation in minimizing cross-region interference and enhancing image fidelity.
\vspace{-10mm} 

\subsubsection{Identifying Hallucination Hotspots in Diffusion Models}
\begin{figure}[H]
     \centering
     \begin{subfigure}[b]{0.55\textwidth}
         \centering
         \includegraphics[width=\textwidth]{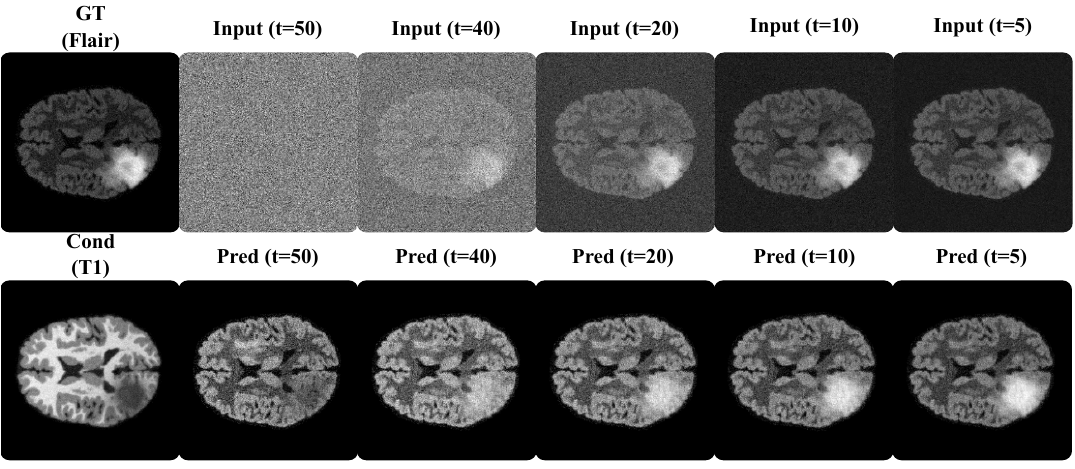}
         \caption{Visual comparison of predictions starting from different intermediate time point.}
         \label{fig:motive3a}
     \end{subfigure}
     \hfill
     \begin{subfigure}[b]{0.43\textwidth}
         \centering
         \includegraphics[width=0.9\textwidth]{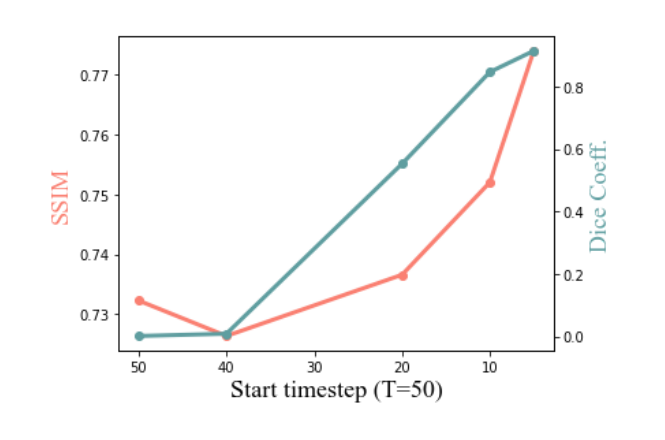}
         \caption{\textit{Dice score} and \textit{SSIM} comparisons}
         \label{fig:motive3b}
     \end{subfigure}
    \caption{Qualitative and quantitative comparisons of predictions starting from different intermediate time points. We sample noisy GT (Flair) and perform a reverse process from it by conditioning the corresponding T1 image.}
    \label{fig:motive3}
\end{figure}

This experiment probes into which stages of the diffusion models' reverse process are most prone to generating hallucinations. Drawing on insights from \cite{sdedit, perceptiondiff}, which noted that diffusion models initially prioritize low-frequency elements before shifting focus to high-frequency details, we explored the hypothesis that certain stages act as hallucination hotspots. Here, we began the reverse process at various intermediate points, introducing Gaussian noise to the ground truth images instead of starting from isotropic Gaussian noise, as shown in Fig~\ref{fig:motive3a}. The findings underscored that hallucinations predominantly emerge during the early to mid-stages, while the final stage is relatively unaffected. Further, as depicted in Fig~\ref{fig:motive3b}, initiating the process closer to the end not only enhances the overall image quality but also better preserves the structural accuracy of features, such as tumors, confirmed by improved dice coefficient and \textit{SSIM} scores.

Our experiments highlight the critical role of OOD-based local image generation and accurate OOD segmentation in reducing hallucinations in diffusion models. By precisely identifying and isolating OOD regions, we minimize cross-region interference, resulting in more accurate image predictions. Additionally, our analysis of the diffusion models' reverse process reveals that the early to mid-stages are particularly vulnerable to hallucination generation.

\subsection{Proposed Framework}

\begin{figure}[!htbp]
\centering
\includegraphics[width=\textwidth]{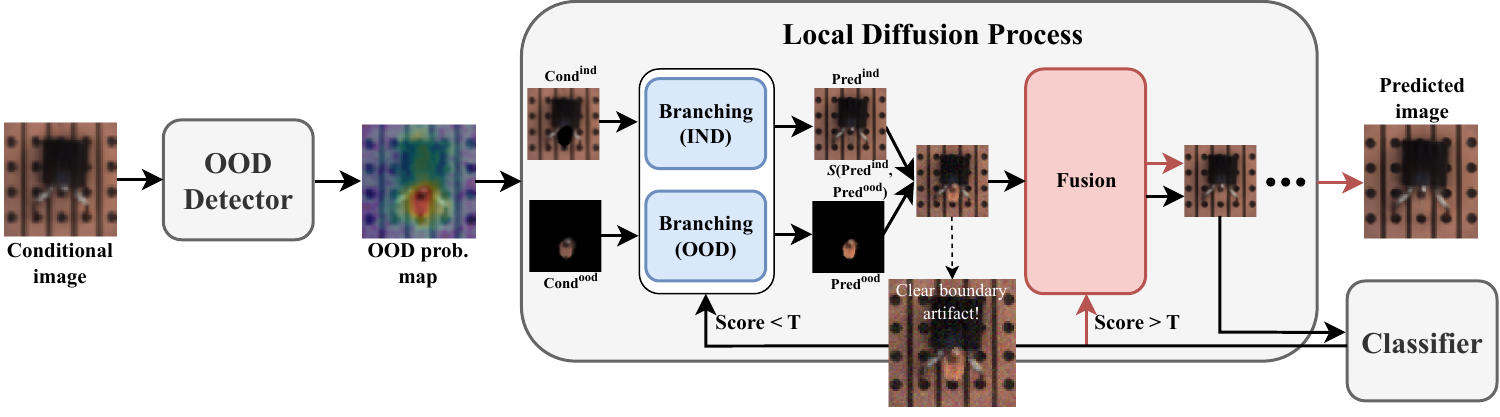}
\caption{Schematic diagram of our proposed framework.}\label{fig:architecture}
\end{figure}

Building on our preliminary findings, we propose a novel framework designed to mitigate structural hallucinations without fine-tuning, as depicted in Fig~\ref{fig:architecture}. This framework employs an OOD detection model to identify OOD regions in conditional images, which are then processed through our \textit{Local Diffusion}, comprising ``branching'' and ``fusion'' modules. This section details the integration and function of these components.

\subsubsection{OOD Estimation}
Our motivational study highlighted the importance of accurately identifying OOD regions in conditional images to enhance our method's efficacy. Should the segmented image inadvertently include a substantial amount of in-distribution (IND) features, the prediction within the OOD region could be adversely influenced, potentially leading to the generation of hallucinated features. Initially, we delineated these regions manually, a process both laborious and time-intensive. To improve efficiency, we adopted PatchCore\cite{patchcore}, a one-class classification model skilled in identifying and segmenting OOD regions. This method utilizes ImageNet pre-trained model embeddings to form a memory bank of features from in-distribution training samples, specifically using WideResNet-50\cite{wideresnet} in our work. During inference, test image embeddings are matched against the memory bank to compute anomalies. Despite the availability of other OOD models, PatchCore\cite{patchcore} was selected for its no-training-required efficiency. In the absence of detectable OOD regions, we revert to the standard reverse process for reliable IND data handling.

\vspace{-10mm} 

\subsubsection{Branching Module}
Upon identifying the OOD regions via a probability map, we generate two masked conditional images, $cond^{ood}$ and $cond^{ind}$, by multiplying the full conditional image by the OOD and IND probabilities respectively. While hard thresholding of the map ensures more precise segmentation, reducing the influence of adjacent regions on the prediction, soft thresholding retains more context about the surrounding areas, enhancing the model's ability to reconstruct the overall image intensity more accurately.

The segmented conditional images, $cond^{ood}$ and $cond^{ind}$, are then processed in parallel to generate separate predictions for OOD ($\hat{x}_{\theta}^{ood}$) and IND ($\hat{x}_{\theta}^{ind}$) regions. Selective masking, based on the OOD map, further refines these predictions. Our approach demonstrates that masking OOD predictions while keeping IND predictions unmasked optimizes integration, preventing the adverse effects of potential inaccuracies in OOD content.

Subsequently, we compute posterior distributions for both OOD and IND regions at each time step \(t\), utilizing shared parameters \(\alpha_{t}\) and \(\beta_{t}\), to accurately model the posterior distribution for each region.
\begin{equation}
\label{eq:posterior_branch}
\begin{gathered}
p(x_{t-1}^{\text{region}} | x_{t}^{\text{region}}) = \mathcal{N}(x_{t-1}^{\text{region}};\mu_{\theta}(x_{t}^{\text{region}}; t), \tilde{\beta_{t}}I)\:\: where \\
\mu_{\theta}(x_{t}^{\text{region}}; t) := \frac{\sqrt{\bar{\alpha}_{t-1}}\beta_{t}}{1-\bar{\alpha}_{t}}\hat{x}_{\theta}^{\text{region}} + \frac{\sqrt{\alpha_{t}}(1-\bar{\alpha}_{t-1})}{1-\bar{\alpha}_{t}}x_{t}^{\text{region}}, \\ \tilde{\beta_{t}} := \frac{1-\bar{\alpha}_{t-1}}{1-\bar{\alpha}_{t}}\beta_{t}.
\end{gathered}
\end{equation}

\subsubsection{Fusion Module}
While the outcomes from branching module lead to the generation of images that are more faithful and exhibit fewer hallucinations, a distinct boundary artifact arises from the local image generation process, as illustrated in Fig~\ref{fig:architecture}. Employing techniques mentioned above such as soft-thresholding helps to reduce this artifact, yet the merged image often appears less natural than those produced by the conventional reverse process. To address this issue, we introduce a fusion module.
\begin{equation}
\label{eq:sum}
\begin{gathered}
S(X_i, Y_i) = 
\begin{cases} 
Y_i & \text{if } cond_{i}^{ood} = 0 \\
X_i & \text{otherwise}
\end{cases}
\:\: where\:\: i \:\:is \:\:each\:\:pixel.
\end{gathered}
\end{equation}
In this phase, the predictions for the OOD and IND regions from the previous branching stage, $\hat{x}_{\theta}^{ood}$ and $\hat{x}_{\theta}^{ind}$, are merged with the noisy inputs corresponding to time step $t$, $x_{t}^{ood}$ and $x_{t}^{ind}$ using the $S$ operator in Equ. \ref{eq:sum}. Consequently, the posterior distribution is formulated as follows:
\begin{equation}
\label{eq:posterior_fusion}
\begin{gathered}
p(x_{t-1} |x_{t}) = \mathcal{N}(x_{t-1}; \mu_{\theta}(x_{t}^{ood}, x_{t}^{ind}; t), \tilde{\beta_{t}}I)\:\: where \\
\mu_{\theta}(x_{t}^{ood}, x_{t}^{ind}; t) := \frac{\sqrt{\bar{\alpha}_{t-1}}\beta_{t}}{1-\bar{\alpha}_{t}}S(\hat{x}_{\theta}^{ood}, \hat{x}_{\theta}^{ind})  + \frac{\sqrt{\alpha_{t}}(1-\bar{\alpha}_{t-1})}{1-\bar{\alpha}_{t}}S(x_{t}^{ood}, x_{t}^{ind}).
\end{gathered}
\end{equation}
Since predicted images for OOD and IND are now combined, the reverse process becomes equivalent to the original DDPM sampling.

\subsubsection{Integration}
To achieve both fidelity and realism in generated images, correctly sequencing the integration of two modules during the reverse process is crucial, as it impacts the balance between these qualities.

Our initial experiments showed that starting the reverse process at an intermediate time step, especially in the later stages reduce hallucinated features. This effectiveness is attributed to the model's focus on refining the overall structure and diminishing residual noise in the latter stages. Based on these findings, we apply ``branching'' module early on, ceasing at a specific intermediate step to avert hallucination. ``Fusion'' module is subsequently introduced in the final stages to improve overall image quality.

Identifying the optimal point for transitioning from ``branching'' to ``fusion'' poses a challenge. We address this with an auxiliary classifier that evaluates the image's classification score $s$ after the first fusion iteration. If $s$ exceeds a set threshold, the fusion proceeds until $t=0$. If not, an additional branching phase precedes further fusion, reassessing the score. The procedure of \textit{Local Diffusion} is detailed in Algo. \ref{alg:pesudo_diffusion}.

\begin{algorithm*}
\caption{Local Diffusion Sampling Scheme}\label{alg:pesudo_diffusion}
\begin{algorithmic}[1]
\scriptsize
\Require pre-trained diffusion model \(f_{\theta}\), classifier \(g_{\phi}\), OOD detector \(h_{\rho}\), condition \(cond\)
\State Initialize  \(t_{intermediate}\), \(x_{T} \sim \mathcal{N}(0, I)\) and compute \(OOD_{prob} \) using \(h_{\rho}(cond)\)
\State Split \(cond\) into \(cond^{ood}\) and \(cond^{ind}\) based on \(OOD_{prob}\)
\State \(flag_{fuse}, flag_{cls} \gets 0, 0\)
\For{\(t = T\) to \(1\)}
    \If{\(t > t_{intermediate}\)} \Comment{Branching for OOD/IND}
        \State Separate \(x_{t}\) into OOD and IND components: \(x_{t}^{ood}\) and \(x_{t}^{ind}\)
        \State Apply posterior sampling separately to OOD and IND prediction parts
    \ElsIf{\(t < t_{intermediate} \) and \(flag_{fuse} == 1\)} \Comment{Fusion module}
        \State Apply posterior sampling with \(\hat{x}_{\theta}\) and \(x_{t}\)
        \If{\(t > 1\) and \(flag_{cls} == 0\)} \Comment{Evaluate with classifier if not last step}
            \State \(s \gets g_{\phi}(\hat{x}_{\theta})\)
            \State \(flag_{fuse}, flag_{cls} \gets s > threshold\) \Comment{Set flags for branching and classifier}
        \EndIf
    \Else
        \State Merge OOD and IND predictions and apply posterior sampling
        \State \(flag_{fuse} = 1\)
    \EndIf
\EndFor
\end{algorithmic}
\end{algorithm*}

\section{Experiments}
\subsection{Experimental Setup}
\subsubsection{Datasets}
We evaluated our method using MNIST ($28\times28$), BraTS\cite{brats} ($224\times224$), and MVTec AD\cite{mvtec} ($224\times224$) datasets. Our experiments involved 2x super-resolution on MNIST and MVTec (anomaly detection), and image translation (T1 to Flair and vice versa) on BraTS (MRI tumor segmentation). To have a meaningful comparison, we evaluated classes that contain structural hallucination examples in the predictions for each dataset. Please see supplementary for further details.

\subsubsection{Implementation Details}
We utilized conditional DDPM with 500 time steps, trained using Adam optimizer (initial learning rate: 0.0001) on NVIDIA A6000 GPUs until model convergence. To address image saturation in RGB super-resolution, we applied interpolation with low-resolution images in the OOD regions up to the fusion stage. See supplementary for further details.
\vspace{-2mm} 

\subsubsection{Evaluation Metrics}
Our evaluation employed the peak signal-to-noise ratio (\textit{PSNR}) and the structural similarity index measure (\textit{SSIM}) for overall image quality assessment but introduced additional downstream tasks to evaluate hallucinations. For MNIST and MVTec, \textit{classification accuracy} were measured, while for BraTS, we used the \textit{dice coefficient} from tumor segmentation. 
\vspace{-2mm} 

\subsection{Main Results}
Tab~\ref{table:results1} shows a quantitative comparison in overall image quality (\textit{PSNR} and \textit{SSIM}) of our approach against various baselines, including DSI \cite{inversionshift}, the sole existing method that tackles out-of-distribution data in image translation without extra training on diffusion models. Our method consistently outperforms DDPM \cite{ddpm} across all evaluated tasks and metrics. 

Despite these results, we argue \textbf{downstream task performance is more important in our task to evaluate the level of hallucination} than \textit{SSIM} and \textit{PSNR} because most hallucinated features are small and localized, minimally impacting these metrics. To address this, we conducted three downstream tasks: \textit{digit classification}, \textit{tumor segmentation}, and \textit{anomaly detection}, with quantitative results shown in Tab~\ref{table:results2}. These results demonstrate significant improvements in reducing structural hallucination with our method. 
Notably, our approach \textit{exceeds DDPM's performance by over threefold in BraTS and by more than 25\% in MVTec, effectively minimizing structural hallucination and reducing misdiagnosis risks.} 
\begin{table}
  \scriptsize
  \vspace{-2mm} 
  \caption{Quantitative comparisons of overall image quality across various datasets, where an upward arrow signifies that a higher value is better. $T$ represents the total number of time steps to sample and p-value (0.05 significance) analysis was conducted.}
  \label{results}
  \centering
  \begin{tabular}{ccccccc}
    \toprule
      & \multicolumn{2}{c}{MNIST} &\multicolumn{2}{c}{BraTS}&\multicolumn{2}{c}{MVTec AD} \\
       & \textit{PSNR} ($\uparrow$) & \textit{SSIM} ($\uparrow$) & \textit{PSNR} ($\uparrow$) & \textit{SSIM} ($\uparrow$) & \textit{PSNR} ($\uparrow$) & \textit{SSIM} ($\uparrow$) \\
    \midrule
    DDPM\cite{ddpm} & 20.8$\pm$1.90 & 0.897$\pm$0.04 & 19.5$\pm$1.75 & 0.709$\pm$0.04 & 26.8$\pm$3.10 & 0.839$\pm$0.10\\
    \myrowcolor%
    DDIM\cite{ddim} ($0.1T$) & 20.8$\pm$1.88 & 0.895$\pm$0.04 & 19.8$\pm$1.53 & 0.715$\pm$0.05 & 27.3$\pm$3.00 & 0.844$\pm$0.10 \\
    DDIM\cite{ddim} ($0.5T$) & 20.6$\pm$1.86 & 0.899$\pm$0.04 & 19.7$\pm$1.61 & 0.703$\pm$0.04 & 27.1$\pm$2.98& 0.840$\pm$0.10\\
    \myrowcolor%
    DDPM $+$ DSI\cite{inversionshift} & 20.8$\pm$1.90 & 0.898$\pm$0.04 & 18.7$\pm$1.74 & 0.695$\pm$0.04 & 26.6$\pm$3.40 & 0.815$\pm$0.10\\
    DDIM $+$ Ours ($0.1T$) & \textbf{20.9$\pm$1.87} & 0.897$\pm$0.04 & 20.7$\pm$1.52 & \textbf{0.720$\pm$0.05} & \textbf{27.5$\pm$3.02} & \textbf{0.847$\pm$0.09}\\
    \myrowcolor%
    DDIM $+$ Ours ($0.5T$) & 20.8$\pm$1.82 & \textbf{0.902$\pm$0.03} & 20.9$\pm$1.61 & 0.711$\pm$0.04& 27.4$\pm$2.97 & 0.844$\pm$0.10 \\
    DDPM $+$ Ours & \textbf{20.9$\pm$1.85} & 0.900$\pm$0.04 & \textbf{21.2$\pm$1.74} & \textbf{0.720$\pm$0.03} & 27.0$\pm$3.05 & 0.843$\pm$0.10 \\
    \myrowcolor%
    \hline
    p-value & \color{blue}0.014 & \color{blue}0.016 & \color{blue}$\approx 0$ &  \color{blue}0.02& \color{blue}0.001 & \color{blue}0.004\\
    \bottomrule
  \end{tabular}
  \label{table:results1}
\end{table}
\vspace{-0mm} 
Furthermore, integrating our method with DDIM \cite{ddim} across various time steps shows superior performance to baseline DDIMs in all datasets, demonstrating robustness to diffusion model hyperparameters. We observe reducing the number of time steps in DDIM leads to decreased accuracy. We root this to fewer time steps leading to more abrupt changes in the prediction at each time step in diffusion models, increasing the risk of hallucinated features. Note that, while DSI\cite{inversionshift} exceeds DDPM in downstream tasks, the performances in overall image quality and downstream tasks are all inferior to our approach.
\vspace{-5mm} 
\begin{table}
  \scriptsize
  \caption{Quantitative results on downstream tasks to measure hallucination.}
  \vspace{-2mm} 
  \label{results}
  \centering
  \begin{tabular}{cccc}
    \toprule
     & MNIST &BraTS&MVTec AD\\
       & \textit{Accuracy (\%)} ($\uparrow$) & \textit{Dice Coefficient} ($\uparrow$) & \textit{Accuracy (\%)} ($\uparrow$) \\
    \midrule
    DDPM\cite{ddpm} & 95.7$\pm$0.61 & 0.194$\pm$0.10 & 58.4$\pm$19.9\\
    \myrowcolor%
    DDIM\cite{ddim} ($0.1T$) & 95.8$\pm$0.61 & 0.256$\pm$0.11  & 53.9$\pm$19.5 \\
    
    DDIM\cite{ddim} ($0.5T$) & 95.9$\pm$0.60 & 0.246$\pm$0.13 & 59.1$\pm$19.4\\
    
    \myrowcolor%
    DDPM $+$ DSI\cite{inversionshift} & 96.0$\pm$0.50 & 0.100$\pm$0.06 & 60.1$\pm$18.6\\
    
    DDIM $+$ Ours ($0.1T$) & 96.1$\pm$0.30 & 0.447$\pm$0.10 & 66.8$\pm$17.8\\
    \myrowcolor%
    DDIM $+$ Ours ($0.5T$) &  97.1$\pm$0.28 & 0.537$\pm$0.06 & 83.2$\pm$14.3 \\
    
    DDPM $+$ Ours & \textbf{98.0$\pm$0.26} & \textbf{0.590$\pm$0.09} & \textbf{85.0$\pm$13.2}\\
    \myrowcolor%
    \hline
    Avg. gain & \color{blue}$\mathcal{+}$1.5\% & \color{blue}$\mathcal{+}$0.293 & \color{blue}$\mathcal{+}$21.2\%\\
    \bottomrule
  \end{tabular}
  \label{table:results2}
\end{table}
\vspace{-5mm} 

\begin{figure}[!htp]
  \centering
  \includegraphics[width=1\textwidth]{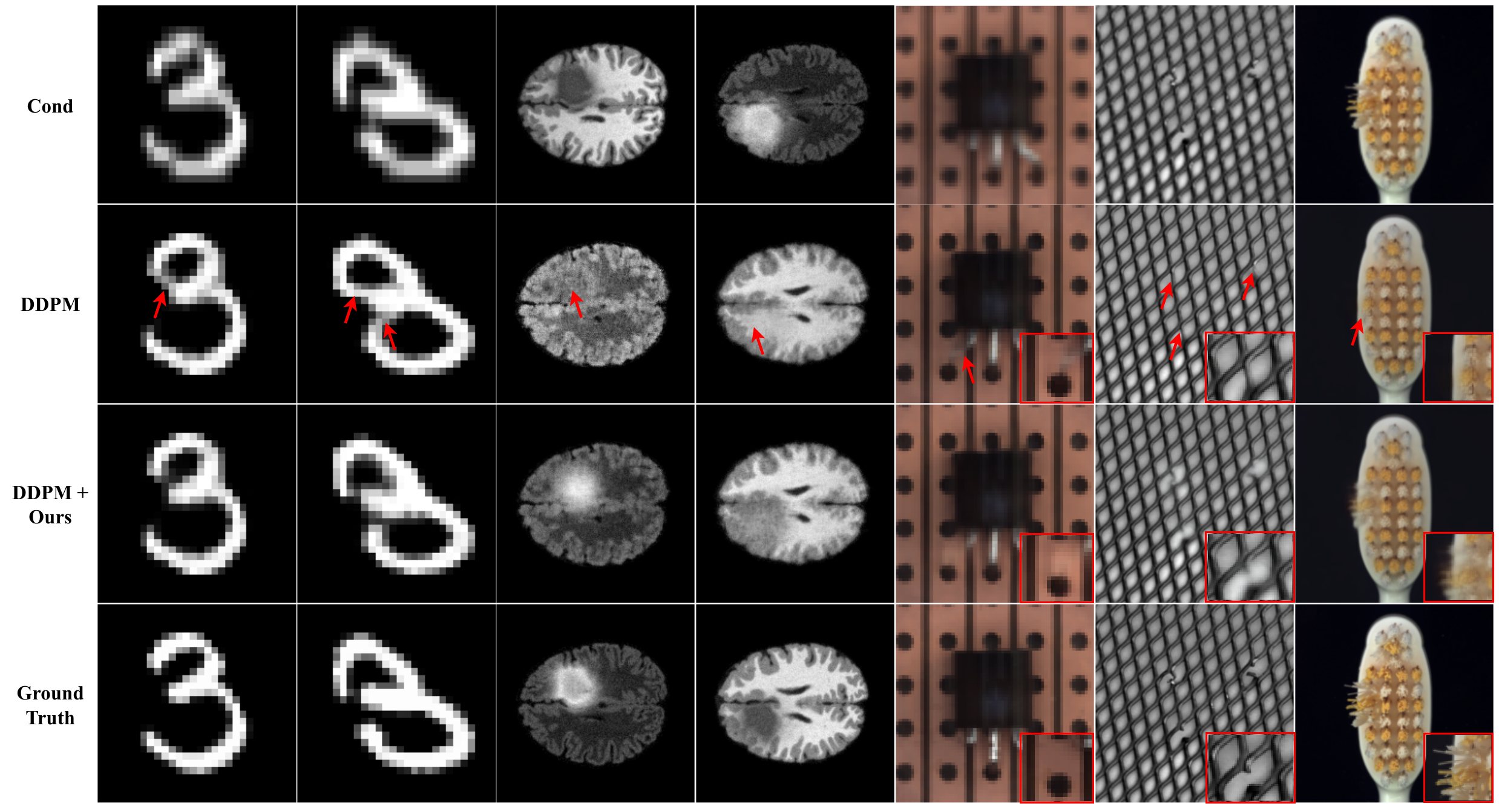}
  \caption{Qualitative comparison on MNIST, BraTS and MVTec (From top: condition, DDPM\cite{ddpm}, DDPM with ours and ground truth). The red arrows indicate structural hallucination.}
  \label{fig:results1}
\end{figure}

Fig~\ref{fig:results1} presents a qualitative comparison against DDPM\cite{ddpm}, showcasing the effectiveness of our proposed method in reducing structural hallucinations. In the BraTS, predictions generated using DDPM\cite{ddpm} contain hallucinated features, notably misidentifying tumor tissues as healthy white matter. In contrast, our method accurately delineates tumors with appropriate intensity ranges. While the MNIST and BraTS feature relatively large OOD areas, the MVTec dataset comprises different types of OOD with various sizes (e.g., broken leads or grids), posing a challenge for hallucination mitigation. Despite these challenges, our approach demonstrates robust performance on various OOD types, as shown in Fig~\ref{fig:results1}, affirming its effectiveness on various image modalities.
\vspace{-2mm} 

\subsection{Further Analysis}
\subsubsection{Performance in OOD/IND Regions}
\begin{figure}[!htbp]
    \centering
    \begin{subfigure}[h]{\textwidth}
        \centering
        \includegraphics[height=3.0cm]{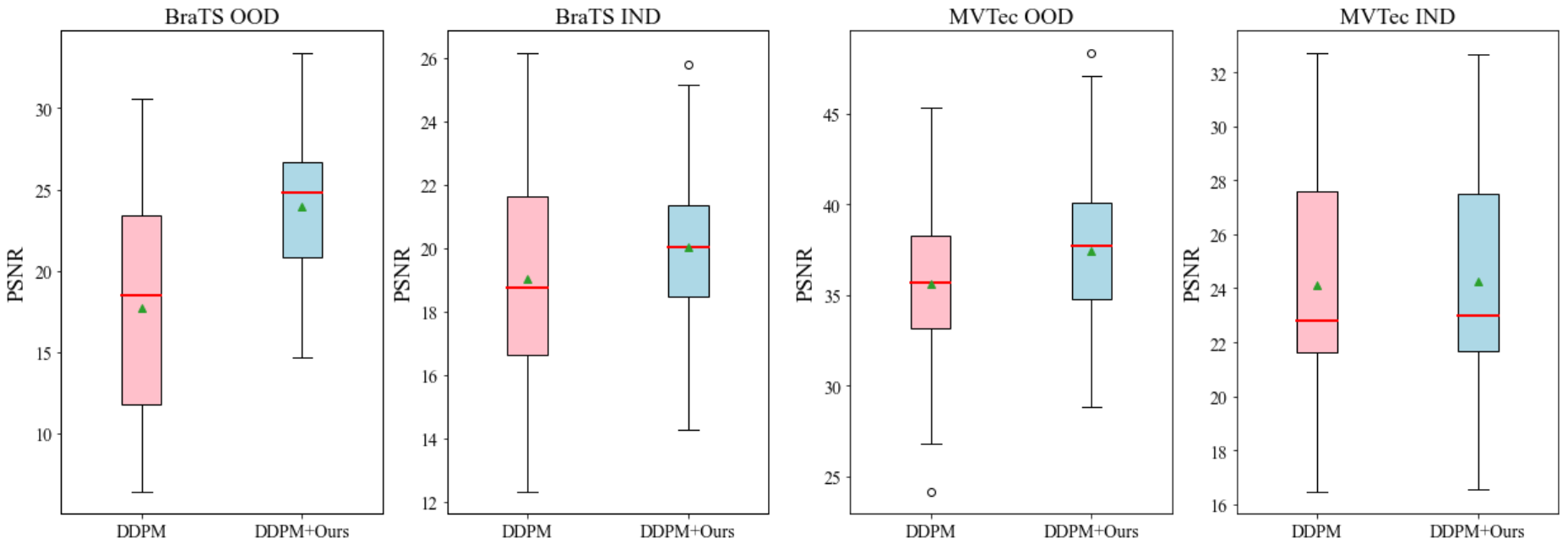}
        \caption{Comparative analysis of performance across individual OOD/IND regions, The red lines and green dots represent the median and mean of each box, respectively.}
        \label{fig:results_separate}
    \end{subfigure}
    \vfill
    \centering
    \begin{subfigure}[h]{\textwidth}
        \centering
        \includegraphics[height=3.0cm]{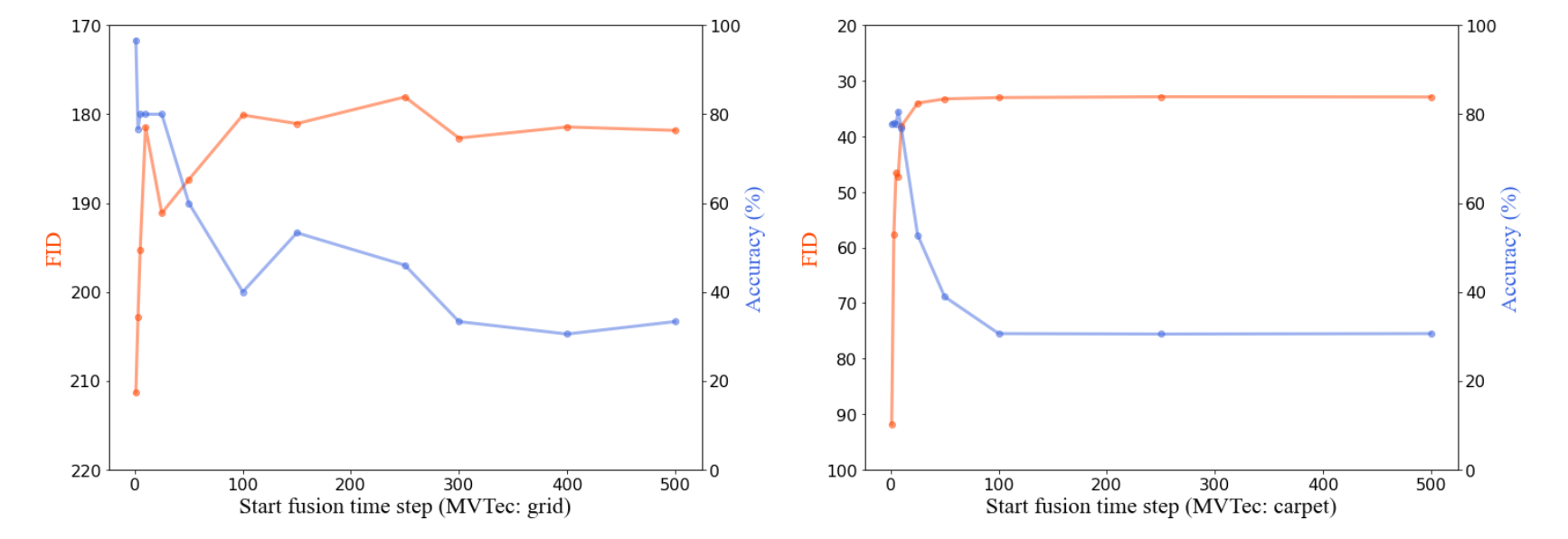}
        \caption{Effect of starting ``fusion'' at various time points in MVTec}
        \label{fig:results_fusion}
    \end{subfigure}
    \vspace{-2mm} 
    \caption{Extensive evaluation on OOD/IND regional performance and the impact of ``fusion'' at various time points.}
    \label{fig:results_further}
\end{figure}

To understand the source of improved performance, in Fig~\ref{fig:results_separate} we investigate the quantitative results in the IND and OOD regions separately. For this analysis, we used the ground truth masks to segment the regions and calculated the \textit{PSNR} of each region. The findings demonstrate that our method surpasses the baselines in terms of both mean and median, exhibiting smaller variance across all datasets. Notably, our method achieves significant improvements in the OOD regions while performing on par with or better than in the IND regions. This suggests that our \textit{Local Diffusion} method does not negatively impact predictions in the IND regions, despite of local image generations.
\vspace{-6mm} 

\vspace{-0mm} 
\begin{table}[H]
    \scriptsize
    \begin{subtable}[H]{0.48\textwidth}
        \centering
          \begin{tabular}{ccccc}
            \toprule
              - & \multicolumn{2}{c}{Single} & \multicolumn{2}{c}{Multiple}\\
               & DDPM & Ours & DDPM & Ours \\ 
            \midrule
            \textit{PSNR} ($\uparrow$) & 19.2 & 22.7 & 19.1 & 21.4 \\
            \textit{Dice coeff.} ($\uparrow$) & 0.032 & 0.667 & 0.055 & 0.540\\
            \bottomrule
          \end{tabular}
       \caption{Model's performance on single and multiple OOD in a single image}
       \label{tab:multipleood}
    \end{subtable}
    \hfill
    \begin{subtable}[H]{0.48\textwidth}
        \centering
          \begin{tabular}{ccccc}
            \toprule
            - & \multicolumn{2}{c}{Small} & \multicolumn{2}{c}{Large}\\
              & DDPM & Ours & DDPM & Ours \\ 
            \midrule
            \textit{PSNR} ($\uparrow$)& 20.7 &  23.4 & 18.0 & 21.5 \\
            \textit{Dice coeff.} ($\uparrow$) & 0.050  &  0.685 & 0.030 & 0.580 \\
            \bottomrule
          \end{tabular}
       \caption{Model's performance on small ($<1.5\%$) and large OOD ($>3\%$) regions}
       \label{tab:largeood}
    \end{subtable}
    \vfill
    \centering
    \begin{subtable}[H]{0.75\textwidth}
      \centering
      \scriptsize
      \begin{tabular}{cccc}
        \toprule
          - & Manual & U-Net Seg.  & PatchCore\\
        \midrule
        \textit{PSNR} ($\uparrow$)&  \textbf{21.8} & \textbf{21.8}  & 21.2 \\
        \textit{Dice coeff.} ($\uparrow$)&  0.670 & \textbf{0.671}  & 0.590 \\
        \bottomrule
      \end{tabular}
      \caption{Comparison of our model's performance using different OOD detectors.}
        \label{table:ood_detector}
    \end{subtable}
    \vspace{-2mm} 
     \caption{Quantitative comparisons on (a,b) various types of OOD and (c) different OOD detectors in BraTS. Visual results are available in the supplementary.}
     \label{tab:ood}
\end{table}
\vspace{-14mm} 

\subsubsection{Balancing Realism and Faithfulness}
A pivotal dilemma within our framework involves determining the ideal moment to transition from the branching stage to the fusion stage. Fig~\ref{fig:results_fusion} demonstrates the consequences of initiating the fusion stage at varying time points evaluated using two metrics: accuracy for faithfulness and Fréchet Inception Distance (FID) for realism. It reveals that commencing the fusion stage towards the end of the process yields higher classification accuracy and FID. In contrast, starting the fusion stage earlier decreases accuracy while improving the FID score. This underscores the inherent trade-off between the realism and faithfulness of our generated images, \textit{highlighting the importance of pinpointing an optimal time point to balance these two aspects}.
\vspace{-3mm} 

\subsubsection{Robustness on different types of OOD} 
Tab~\ref{tab:multipleood} and \ref{tab:largeood} quantitatively compare our model's performance across different OOD scenarios, such as \textit{varying sizes} and \textit{multiple OOD occurrences}. For this, we employ annotated labels for segmenting OOD instances to ensure accurate evaluation of our \textit{Local Diffusion}. 
As demonstrated in Tab~\ref{tab:multipleood}, although the presence of multiple OOD instances within a single image introduces extra complexity and leads to a decrease in the \textit{Dice score}, our approach still significantly enhances performance across both metrics compared to the baseline. Tab~\ref{tab:largeood} further illustrates our model's ability to handle OOD of various sizes, underscoring that \textit{Local Diffusion} effectively improves upon the baseline across different OOD dimensions. Overall, our \textit{Local Diffusion} strategy is an effective tool for reducing hallucination effects in images, accommodating a wide range of sizes and occurrences of OOD.
\vspace{-5mm} 

\subsubsection{Impact of OOD Estimation Accuracy}
Our framework relies on the initial estimation of OOD region from conditional images, a process critical for accurate segmentation, as illustrated in Fig~\ref{fig:motive2}. To underscore the importance of this step, we assessed the quality of generated images using various OOD detectors, including manual segmentation, U-Net segmentation \cite{unet}, and PatchCore\cite{patchcore} with WideResNet50\cite{wideresnet}. The results, detailed in Tab~\ref{table:ood_detector}, reveal a significant performance increase when manual or U-Net segmentation is employed instead of PatchCore. This outcome underscores the pivotal role that precise segmentation of the OOD region plays in mitigating hallucinations, marking it as a critical focus for future research endeavors. We provide more comprehensive comparisons of different OOD estimation methods in the supplementary section.
\vspace{-5mm} 

\begin{figure}[!htbp]
\centering
\includegraphics[width=0.9\textwidth]{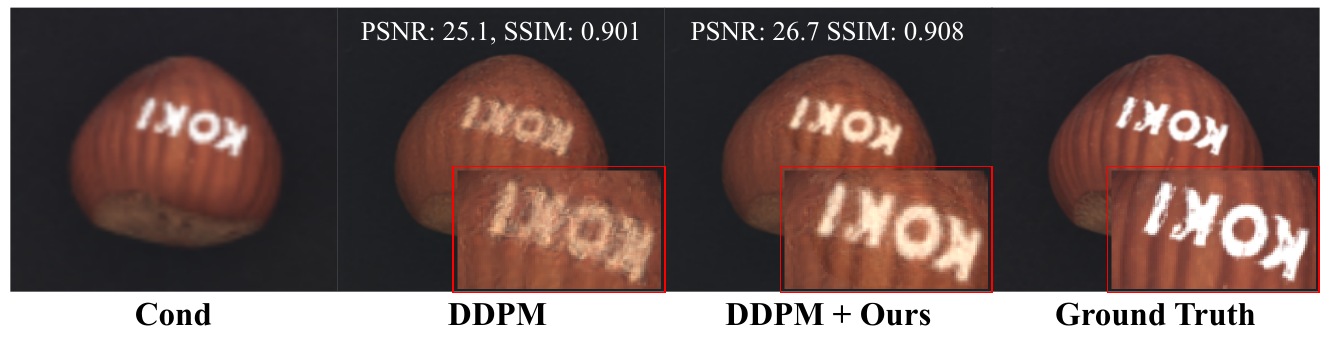}
\vspace{-2mm} 
\caption{Visual comparison of DDPM and ours on color hallucination. While ours mitigates structural hallucination, DDPM fails to predict RGB values in the OOD region.}
\label{fig:colorhallu}
\end{figure} 
\vspace{-11mm} 

\section{Conclusions}
\vspace{-1mm} 

We introduce a novel framework to reduce structural hallucinations in conditional diffusion models. Through motivational experiments, we validated our hypothesis that localized image generation markedly diminishes structural hallucinations. We discovered that diffusion models mainly produce hallucinations during the early to mid phases of the reverse process. Based on these insights, we developed a ``branching'' module for localized generation via OOD/IND segmentation, and a ``fusion'' module to merge these local predictions into cohesive, natural images seamlessly. These modules are designed for easy integration with any diffusion model without additional training, offering an efficient way to mitigate hallucinations.

Our evaluation across multiple image datasets demonstrated significant improvements in image quality and fidelity, with visual assessments confirming reduced hallucinations, significantly lowering the risk of misdiagnosis. These results were consistent across different diffusion models and image resolutions.

Future efforts could focus on more efficient and interpretable method for OOD detection and broaden the evaluation to more hallucination types such as color or texture, as suggested in Fig~\ref{fig:colorhallu}. We anticipate our work will inspire further exploration into counteracting hallucinations in diffusion models for diverse image translation applications, crucial for fields such as medical imaging and automated manufacturing.
\vspace{-3mm} 

\section*{Acknolwedgement}
\vspace{-2mm} 

This work was partially supported by an AstraZeneca-funded internship and the EPSRC-funded UCL Centre for Doctoral Training in Intelligent, Integrated Imaging in Healthcare (i4health) under grant number EP/S021930/1. Wellcome Trust award 221915/Z/20/Z, MRC award MR/W031566/1 and the NIHR UCLH Biomedical Research Centre support the work of DCA, MF, and HG on this topic.

%
%
\bibliographystyle{splncs04}
\bibliography{egbib}

\newpage
\appendix
\renewcommand{\thesection}{\Alph{section}}
\renewcommand{\thesubsection}{\thesection.\arabic{subsection}}
\section{Supplementary Material}
\subsection{Qualitative Results}
\begin{figure}[H]
  \centering
  \includegraphics[width=0.95\textwidth]{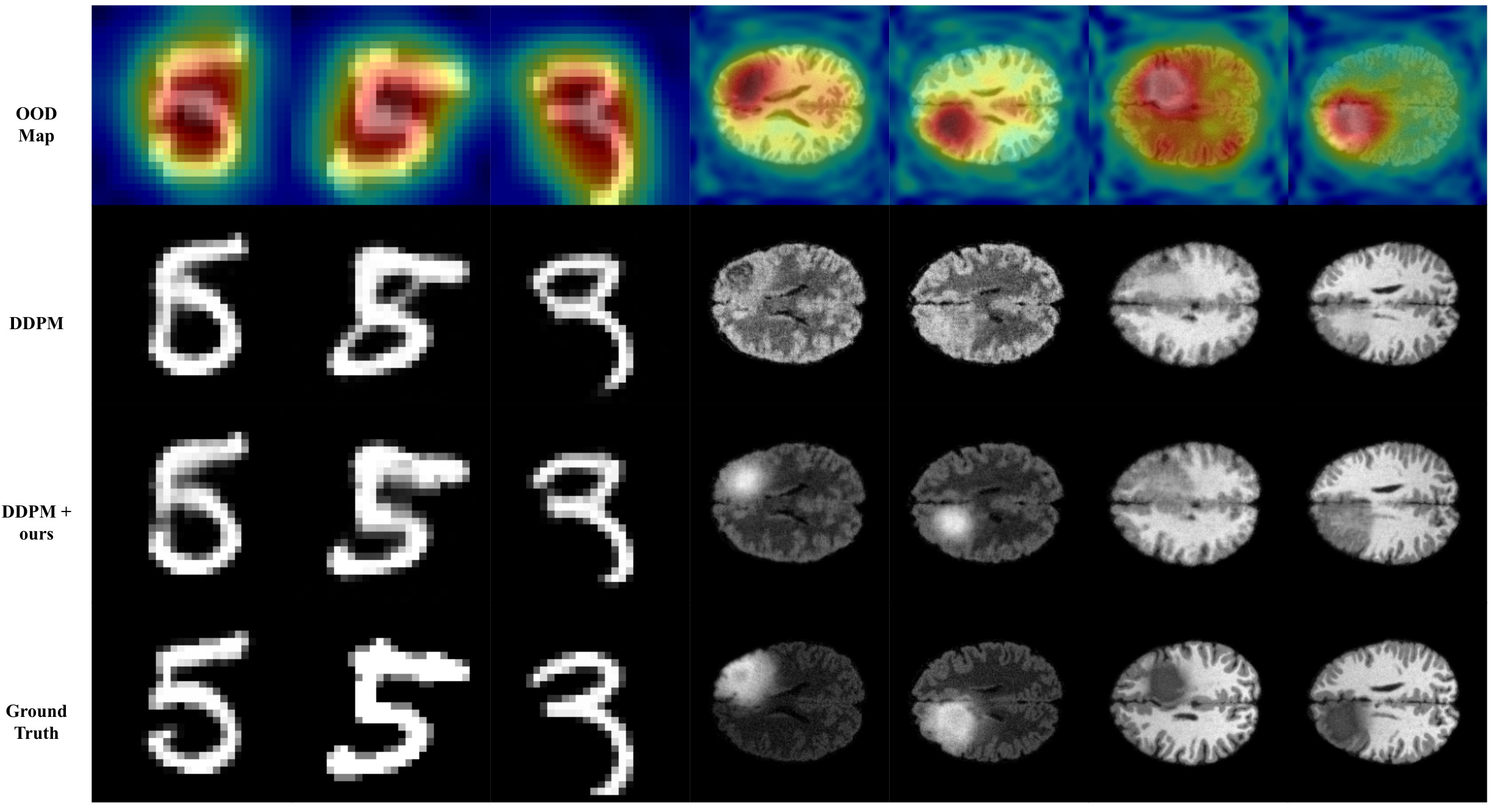}
  \caption{Qualitative comparison on MNIST and BraTS(From top: Predicted OOD map, DDPM\cite{ddpm}, DDPM with ours and ground truth).}
  \label{fig:supple1}
\end{figure}

\begin{figure}[H]
  \centering
  \includegraphics[width=0.95\textwidth]{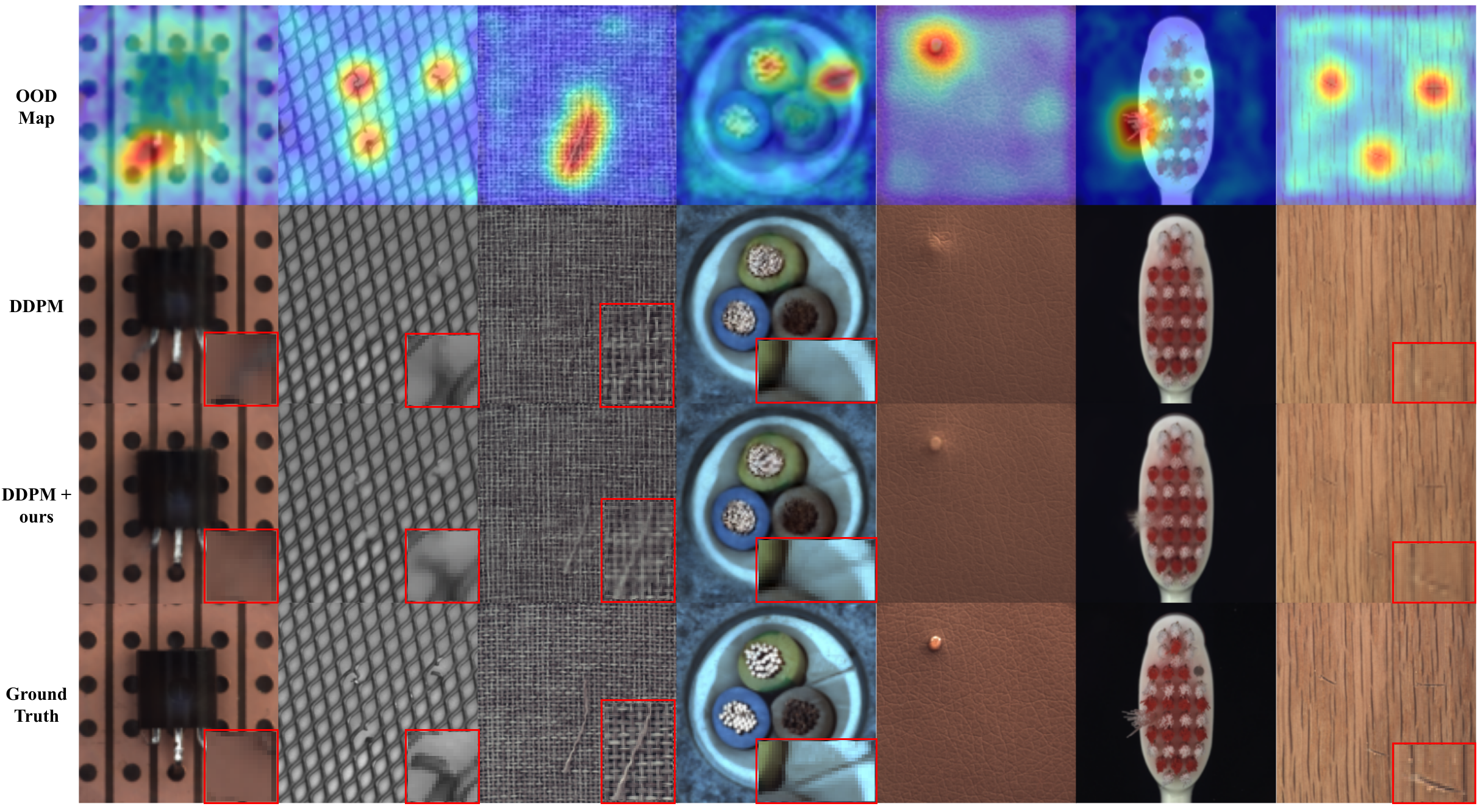}
  \caption{Qualitative comparison on MVTec AD (From top: Predicted OOD map, DDPM\cite{ddpm}, DDPM with ours and ground truth).}
  \label{fig:supple2}
\end{figure}

Fig~\ref{fig:supple1} and \ref{fig:supple2} present visual comparisons across the MNIST, BraTS, and MVTec AD datasets. We conducted image super-resolution on MNIST and MVTec AD, and image translation on BraTS. The first rows in each figure display the predicted OOD maps of conditional images using our OOD detector, the second rows showcase hallucinated predictions from the DDPM model, and the third rows depict predictions produced by our \textit{Local Diffusion} approach. 
\vspace{-5mm} 
\vspace{-15mm} 
The results demonstrate the efficacy of our \textit{Local Diffusion} across different image modalities. As shown in the penultimate column's example in Fig~\ref{fig:supple1}, our method maintains robust performance even in the presence of a noisy OOD probability map, alleviating the need for precise OOD region estimation and avoiding false positives.

\begin{figure}[H]
  \centering
  \includegraphics[width=0.9\textwidth]{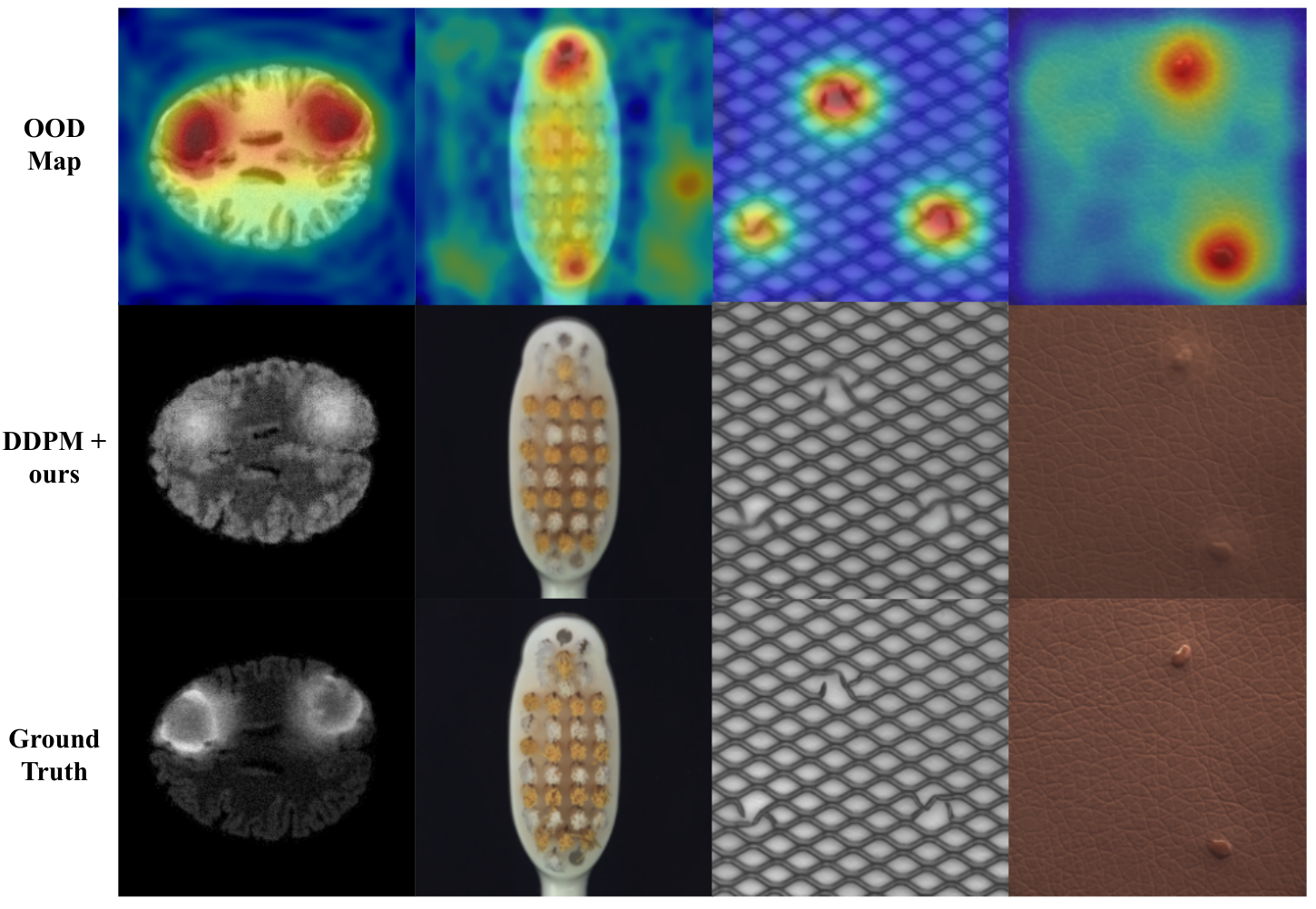}
  \caption{\textbf{Our method shows robustness with varying size and occurrences of OOD regions in one image}: visualization of various OOD cases. (From top: Predicted OOD map, DDPM with ours and ground truth).}
  \label{fig:supple3}
\end{figure}

Fig~\ref{fig:supple3} displays the qualitative outcomes of applying our \textit{Local Diffusion} method across diverse OOD scenarios. Each test image encompasses a range of OOD types, including variations in the size and frequency of OOD regions. Despite these variations, our approach consistently demonstrates robust performance, effectively restoring OOD regions while avoiding hallucinations.

\subsection{Quantitative Results}
\begin{figure}[H]
  \centering
  \includegraphics[width=0.93\textwidth]{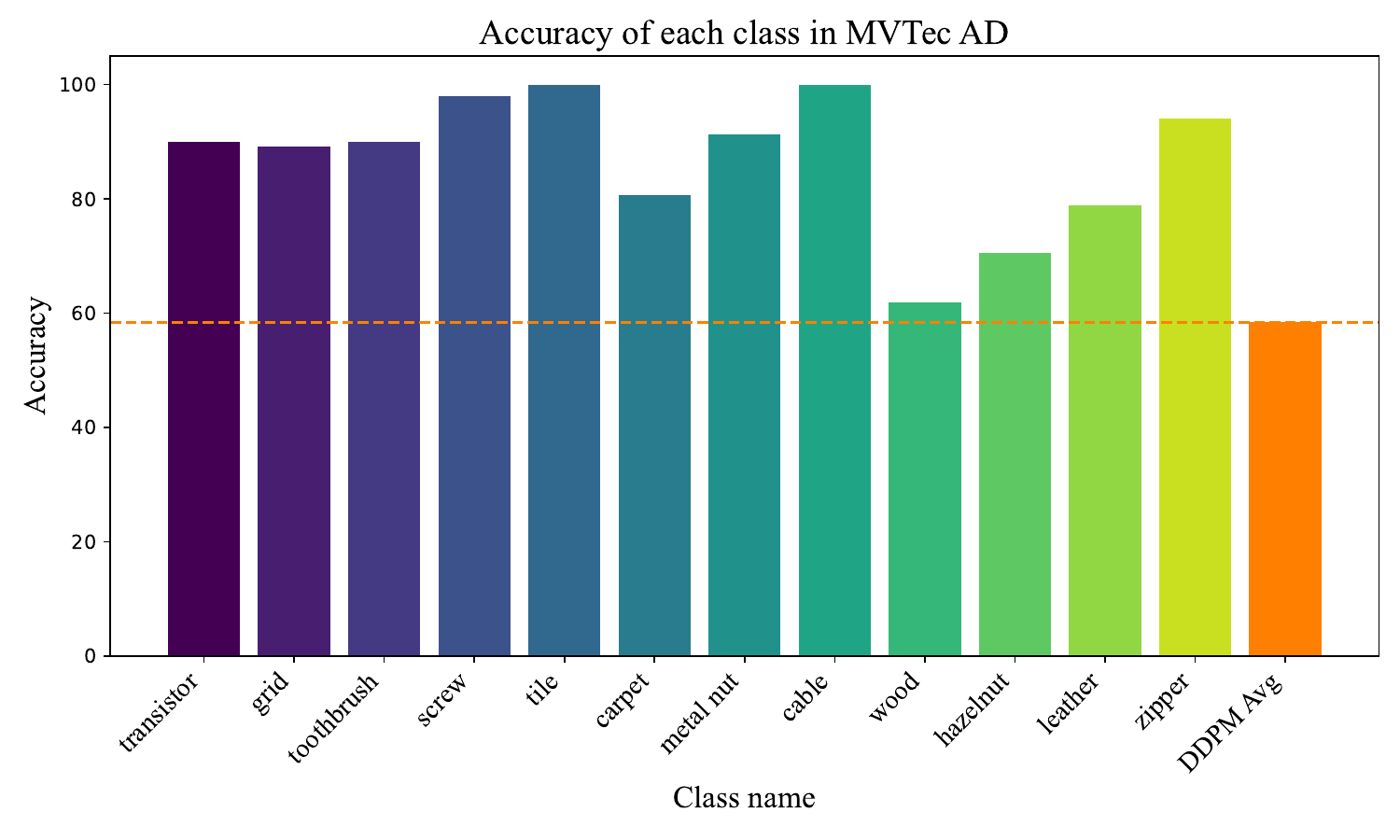}
  \caption{Quantitative results of our approach across each object in MVTec AD.}
  \label{fig:supple_class}
\end{figure}

Following from Tab~\ref{table:results2}, Fig~\ref{fig:supple_class} illustrates our model's performance on the downstream task for MVTec AD across each class. For this task, we trained an anomaly detection model to determine whether the predicted image is anomalous. To highlight the distinctions, we evaluated using only anomalous data. The results demonstrate that our \textit{Local Diffusion} reduces structural hallucination, with each class's classification accuracy exceeding the average accuracy of DDPM (58.4\%).
\vspace{-15mm} 
\begin{figure}[H]
  \centering
  \includegraphics[width=0.93\textwidth]{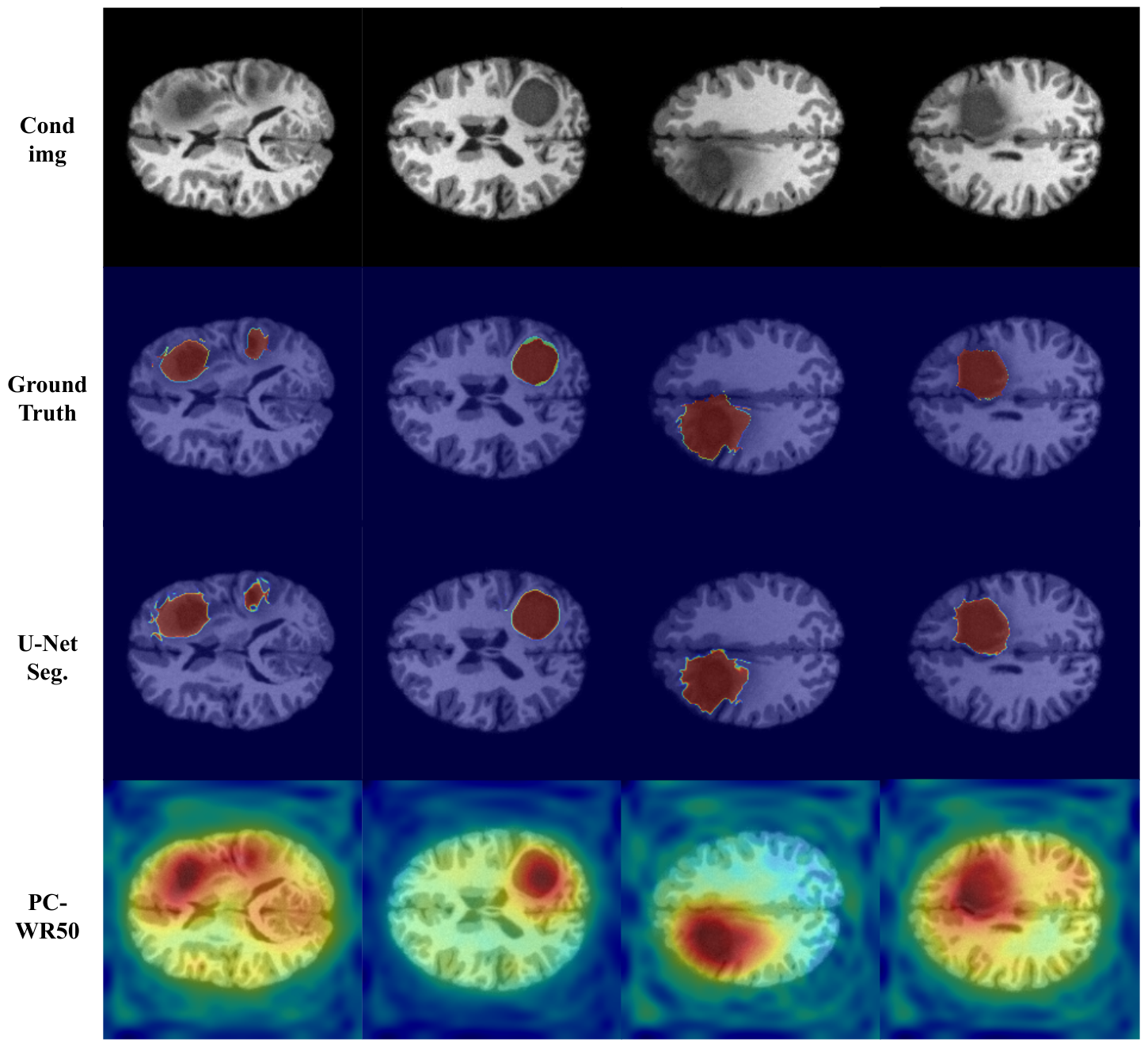}
  \caption{Visual comparison of different OOD detectors in BraTS.}
  \label{fig:ood_detector2}
\end{figure}

\subsubsection{Impact of OOD Detection Method}
Fig~\ref{fig:ood_detector2} shows visual comparisons of PatchCore and a supervised method using U-Net. As can be seen, more precise OOD segmentation can be performed using either ground truth masks or U-Net for OOD region segmentation (e.g., tumors). While this can significantly improves image quality and reduces hallucinations further using our \textit{Local Diffusion}, employing ground truth masks is labor-intensive for each sample, and U-Net segmentation requires expensive training with both IND and OOD samples, making these approaches infeasible in real applications.
Tab~\ref{table:ood_detector2} compares our model's performance using various OOD detection methods. Results show that PatchCore\cite{patchcore} outperforms other detectors in terms of both performance and efficiency. Hence, PatchCore offers the most cost-effective automated OOD detection among the baselines, without requiring OOD samples or expensive back-propagation during training, and is more adaptable to various OOD types.
\vspace{-2mm} 

\begin{table}[!hbpt]
  \centering
  \small
  \begin{tabular}{cccc}
    \toprule
      - & Dice & Train/Inf. time (s) & Gradient descent\\
    \midrule
    CFA\cite{cfa} &  0.698 & 140.8/\textbf{0.017} & Yes  \\
    FastFlow\cite{fastflow} & 0.670 & 166.1/0.032 & Yes \\
    PatchCore &\textbf{0.789} & \textbf{13.2}/\textbf{0.017} & No \\
    \bottomrule
  \end{tabular}
  \caption{Comparison of different OOD detectors on BraTS.}
   \label{table:ood_detector2}
\end{table}
\vspace{-13mm} 

\subsection{Dataset}
\vspace{-1mm} 
We evaluated our method on the MNIST, BraTS, and MVTec AD datasets, performing 2x super-resolution for MNIST and MVTec (anomaly detection) and image translation for BraTS (MRI tumor segmentation). MNIST includes 10,000 test images, while BraTS has 92 3D MR scans, from which we extracted 4,419 2D slices (indexes 70-110) in axial view across T1, T2, and Flair modalities. MVTec AD contains 15 sub-datasets with 5,354 images in total, 1,725 of which are test images featuring normal and defective samples for various products, complete with anomaly masks.

For a comparative analysis, we focused on classes showing \textit{structural hallucinations} upon visual inspection. All BraTS MR scans were used due to tumor presence. In MVTec AD, we selected 12 classes exhibiting hallucinations, excluding 'bottle', 'capsule' and 'pill'. Quantitative analyses in MVTec AD excluded nominal class samples.
\vspace{-3mm} 

\subsection{Implementation Details}
\subsubsection{OOD Detector - PatchCore}
For OOD detection, PatchCore was employed, utilizing WideResNet-50 as its backbone. Embeddings were extracted from the first two layers of the ResNet block. The coreset sampling ratio was set to 0.1 for MNIST and MVTec AD, and to 0.01 for BraTS.
\vspace{-2mm} 

\subsubsection{Downstream Task - Digit Classification}
For MNIST downstream task evaluation, we trained a 5-layer (2 convolutional, 1 pooling, and 2 fully connected) neural network with ReLU activation and base dimension of 32 channels. 
\vspace{-2mm} 

\subsubsection{Downstream Task - Tumor Segmentation}
For tumor segmentation in BraTS, we trained a vanilla U-Net with input channel 1. Since this is a short-tail problem, we used weighted cross entropy with dice loss functions to optimize the weights of the U-Net.
\vspace{-2mm} 

\subsubsection{Downstream Task - Anomaly Detection}
For MVTec AD, we also trained a classification model but with a one-class classification approach using PatchCore with WideResNet50 backbone. We trained a single PatchCore using the whole nominal training samples in the dataset.
\vspace{-2mm} 

\begin{table}[H]
  \centering
  \begin{tabular}{cccc}
    \toprule
      Dataset & MNIST & BraTS & MVTec AD\\
    \midrule
    No. of time steps& 500 & 50, 100 & 500\\
    Image size& 28 & 224 & 112, 224\\
    Parametrization & $x$ (target image) & $x$ (target image) & $x$ (target image)\\
    Noise scheduler & Sigmoid & Sigmoid & Sigmoid\\
    Network & Attention U-Net & Attention U-Net & Attention U-Net \\
    Number of params. & 3.35M & 12.1M & 12.1M \\
    $t_{intermediate}$ & 2 & 10 & 5 \\
    \bottomrule
  \end{tabular}
  \caption{Implementation details of DDPM for each dataset}
   \label{table:implementation}
\end{table}

\subsection{Additional Experiments}
\begin{figure}[H]
  \centering
  \includegraphics[width=\textwidth]{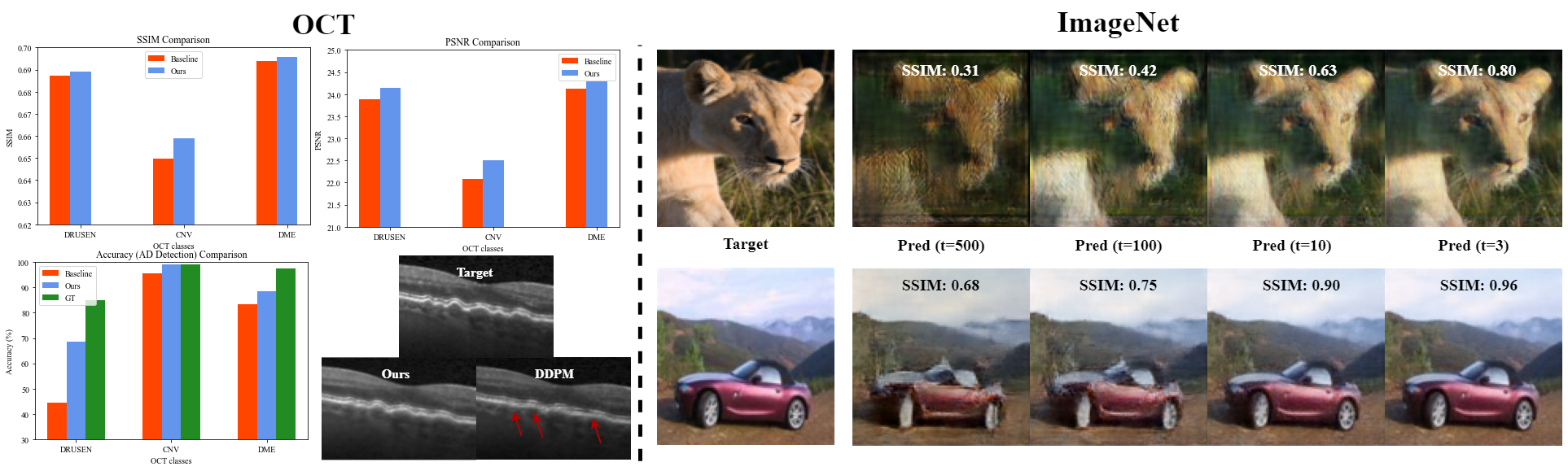}
  \caption{Quantitative and visual comparisons on OCT and ImageNet. $t$ refers to the intermediate starting point.}
  \label{fig:add_exp}
\end{figure}

Here, we present additional results in Fig~\ref{fig:add_exp} for the OCT\cite{oct} (full exp.) and ImageNet\cite{imagenet} (motivational exp.), which are more complex in shapes and colors. \textit{Local Diffusion} consistently outperforms DDPM in both image quality and downstream evaluation for OCT. Fig~\ref{fig:add_exp} also shows visual comparisons with varying starting points in the reverse process, using noisy ground truth (GT) as inputs as done in our motivational study. The models trained on `Tiger' and `Boat', are tested on `Lion' and `Car', respectively. The results show hallucination effects diminish as the starting point moves to the end. This again supports our conclusion `the early to mid-stages are vulnerable to hallucination generation'.

\subsection{Detailed Derivation of Hypothesis}

To further elaborate on the justification, we start by considering the Evidence Lower Bound (ELBO) in the context of the Denoising Diffusion Probabilistic Models (DDPM). The ELBO is a variational bound on the log-likelihood of the data and is defined as:
\begin{equation}
\begin{gathered}
ELBO = \mathbb{E}_{q}[\log p(x_0) - \sum_{t=1}^{T} KL(q(x_{t-1}|x_{t}) \| p(x_{t-1}|x_{t}))]
\end{gathered}
\end{equation}

Using the definition of KL divergence:
\begin{equation}
KL(q(x_{t-1}|x_{t}) \| p(x_{t-1}|x_{t})) = \int q(x_{t-1}|x_{t}) \log \frac{q(x_{t-1}|x_{t})}{p(x_{t-1}|x_{t})} dx_{t-1}
\end{equation}

When we introduce the partitioning of $x_t$ and $x_{t-1}$ into in-distribution and out-of-distribution components, but both conditioned to the un-partitioned $x_{t}$., the ELBO can be expanded to separately consider these components. For the estimation process, we have:
\begin{equation}
\begin{gathered}
ELBO_{partition}^{est} = \mathbb{E}_{q}[\log p(x_0) - \\ \sum_{t=1}^{T} \left( \int q(x_{t-1}^{in}|x_{t}) \log \frac{q(x_{t-1}^{in}|x_{t})}{p(x_{t-1}^{in}|x_{t})} dx_{t-1}^{in} + \int q(x_{t-1}^{out}|x_{t}) \log \frac{q(x_{t-1}^{out}|x_{t})}{p(x_{t-1}^{out}|x_{t})} dx_{t-1}^{out} \right)]
\end{gathered}
\end{equation}

By conditioning the estimation on the partitioned components $x_t^{in}$ and $x_t^{out}$, the ELBO can be refined as:
\begin{equation}
\scriptsize
\begin{gathered}
ELBO_{partition}^{est\_cond} = \mathbb{E}_{q}[\log p(x_0) - \\ \sum_{t=1}^{T} \left( \int q(x_{t-1}^{in}|x_{t}^{in}) \log \frac{q(x_{t-1}^{in}|x_{t}^{in})}{p(x_{t-1}^{in}|x_{t}^{in})} dx_{t-1}^{in} + \int q(x_{t-1}^{out}|x_{t}^{out}) \log \frac{q(x_{t-1}^{out}|x_{t}^{out})}{p(x_{t-1}^{out}|x_{t}^{out})} dx_{t-1}^{out} \right)]
\end{gathered}
\end{equation}

Given that $x_{t}^{in}$ belongs to $x_{t}$ and is closer to the distribution of $x_{t-1}^{in}$, we can assume that:
\begin{equation}
\scriptsize
\sum_{t=1}^{T} \left( \int q(x_{t-1}^{in}|x_{t}) \log \frac{q(x_{t-1}^{in}|x_{t})}{p(x_{t-1}^{in}|x_{t})} dx_{t-1}^{in} \right) \geq \sum_{t=1}^{T} \left( \int q(x_{t-1}^{in}|x_{t}^{in}) \log \frac{q(x_{t-1}^{in}|x_{t}^{in})}{p(x_{t-1}^{in}|x_{t}^{in})} dx_{t-1}^{in} \right)
\end{equation}

Similarly, for the out-of-distribution components:
\begin{equation}
\scriptsize
\sum_{t=1}^{T} \left( \int q(x_{t-1}^{out}|x_{t}) \log \frac{q(x_{t-1}^{out}|x_{t})}{p(x_{t-1}^{out}|x_{t})} dx_{t-1}^{out} \right) \geq \sum_{t=1}^{T} \left( \int q(x_{t-1}^{out}|x_{t}^{out}) \log \frac{q(x_{t-1}^{out}|x_{t}^{out})}{p(x_{t-1}^{out}|x_{t}^{out})} dx_{t-1}^{out} \right)
\end{equation}

Here, the partitioned estimation and conditioning allow for a more precise calculation of the KL divergence terms, as each term specifically addresses the relevant data distribution. This leads to a reduction in the overall KL divergence, thus potentially increasing the ELBO:
\begin{equation}
\begin{gathered}
ELBO_{partition}^{est} = \mathbb{E}_{q}[\log p(x_0) - \sum_{t=1}^{T} (KL(q(x_{t-1}^{in}|x_{t}) \| p(x_{t-1}^{in}|x_{t})) \\ + KL(q(x_{t-1}^{out}|x_{t}) \| p(x_{t-1}^{out}|x_{t})))] 
 \leq \\ ELBO_{partition}^{est\_cond} = \mathbb{E}_{q}[\log p(x_0) - \sum_{t=1}^{T} (KL(q(x_{t-1}^{in}|x_{t}^{in}) \| p(x_{t-1}^{in}|x_{t}^{in})) \\ + KL(q(x_{t-1}^{out}|x_{t}^{out}) \| p(x_{t-1}^{out}|x_{t}^{out}))]
\end{gathered}
\end{equation}

Given that $ELBO_{in}^{est} \leq ELBO_{in}^{est\_cond}$ and $ELBO_{out}^{est} \leq ELBO_{out}^{est\_cond}$, the total ELBO when partitioned and conditioned appropriately can be expected to be higher or equal to the unpartitioned ELBO. This demonstrates the advantage of our approach in partitioning and conditioning on in-distribution and out-of-distribution components separately for the denoising process.

\end{document}